# Random Feature Maps for Dot Product Kernels

**Purushottam Kar** and **Harish Karnick**
Indian Institute of Technology Kanpur, INDIA
{purushot,hk}@cse.iitk.ac.in

## Abstract

Approximating non-linear kernels using feature maps has gained a lot of interest in recent years due to applications in reducing training and testing times of SVM classifiers and other kernel based learning algorithms. We extend this line of work and present low distortion embeddings for dot product kernels into linear Euclidean spaces. We base our results on a classical result in harmonic analysis characterizing all dot product kernels and use it to define randomized feature maps into explicit low dimensional Euclidean spaces in which the native dot product provides an approximation to the dot product kernel with high confidence.

## 1 Introduction

Kernel methods have gained much importance in machine learning in recent years due to the ease with which they allow algorithms designed to work in linear feature spaces to be applied to implicit non linear feature spaces. Typically these non linear feature spaces are high (often infinite) dimensional and in order to avoid incurring the cost of explicitly working in these spaces, one invokes the well known *kernel trick* which exploits the fact that the algorithms in question interact with data solely through pairwise inner products. For example, instead of directly learning a hyperplane classifier in $\mathbb{R}^d$, one considers a non linear map $\Phi : \mathbb{R}^d \to \mathcal{H}$ such that for all $\mathbf{x}, \mathbf{y} \in \mathbb{R}^d, \langle \Phi(\mathbf{x}), \Phi(\mathbf{y}) \rangle_{\mathcal{H}} = K(\mathbf{x}, \mathbf{y})$ for some easily computable kernel $K$. One then tries to learn a classifier $H : \mathbf{x} \mapsto \mathbf{w}^\top \Phi(\mathbf{x})$ for some $\mathbf{w} \in \mathcal{H}$.

However, one is faced with the problem of representation in these non linear feature spaces and is at the risk of incurring the curse of dimensionality. The solution to this problem comes in the form of *Representer Theorems* (see Argyrioua et al., 2009, for recent results) which act as an implicit dimensionality reduction step by giving us an assurance that the object(s) of interest, for example the normal vector to the hyperplane $\mathbf{w}$ in the case of classification and non-linear regression, the cluster centers in the case of kernel $k$-means, or the principal components in the case of kernel PCA, would necessarily lie in the span of the non-linear feature maps of the training vectors in the respective examples (see Schölkopf and Smola, 2002). For instance, in case of the SVM algorithm, the result ensures that the maximum margin hyperplane in $\mathcal{H}$ would necessarily be of the form $\mathbf{w} = \sum \alpha_i \Phi(\mathbf{x}_i)$ where $\mathbf{x}_i$ are the training points. In case of SVM regression and classification, such a result is arrived at by application of the Karush-Kuhn-Tucker conditions whereas in the other two applications, the respective formulations themselves yield such a result.

Whereas this appears to solve the problem of the curse of dimensionality, it actually paves the way for an entirely new kind of curse – one that we call the *Curse of Support*. In order to evaluate the output of the algorithms on test data, say in the case of SVM classification, one has to compute the kernel measures of the test point with all the training points that participate in defining the normal vector $\mathbf{w}$. This cost can be prohibitive if the support is large. Unfortunately this is almost surely the case with large datasets as demonstrated by several results (Steinwart, 2003, Steinwart and Christmann, 2008, Bengio et al., 2005) which predict an unbounded growth in the support sizes with growing training set sizes. A similar fate awaits all other kernel algorithms that use the support vector effect in order to avoid explicit representations.

This presents a dilemma where a large training set is beneficial in obtaining superior generalization properties but is simultaneously responsible in slowing the algorithms' predictive routines. There has been a lot of research on SVM formulations with sparsity pro-moting regularizers (see for example Bi et al., 2003) and support vector reduction (see for example Cossalter et al., 2011). However, although these efforts have yielded rich empirical returns, they have neither addressed other kernel algorithms nor approached the question behind the curse in a systematic way.

## 2 Related Work

In a very elegant result, Rahimi and Recht (2007) demonstrated how this curse can be beaten by way of low-distortion embeddings. Their result, building upon a classical result in harmonic analysis called Bochner's Theorem (refer to Rudin, 1962), shows how to, in some sense, embed the non-linear feature space (i.e. $\mathcal{H}$, the Reproducing Kernel Hilbert Space associated with the kernel $K$) into a low dimensional Euclidean space while incurring an arbitrarily small additive distortion in the inner product values. More formally they constructed randomized feature maps $Z : \mathbb{R}^d \to \mathbb{R}^D$ such that for $\mathbf{x}, \mathbf{y} \in \mathbb{R}^d, \langle Z(\mathbf{x}), Z(\mathbf{y}) \rangle \approx K(\mathbf{x}, \mathbf{y})$ with very high probability.

This allows one to overcome the curse of support in a systematic way for all the kernel learning tasks mentioned before since one may now work in the explicit low dimensional space $\mathbb{R}^D$ with explicit representations whose complexity depends only on the dimensionality of the space. Their contribution is reminiscent of Indyk and Motwani (1998) who perform low distortion embeddings (by invoking the Johnson-Lindenstrauss Lemma) in order to overcome the curse of dimensionality for the nearest neighbor problem.

Subsequently there has been an increased interest in the kernel learning community toward results that allow one to use linear kernels over some transformed feature space without having to sacrifice the benefits provided by non-linear ones. Rahimi and Recht (2007) considered only translation invariant kernels i.e. kernels of the form $K(\mathbf{x}, \mathbf{y}) = f(\mathbf{x} - \mathbf{y})$ for some positive definite function $f : \mathbb{R}^d \to \mathbb{R}$. Subsequently Li et al. (2010) generalized this to a larger class of group invariant kernels while still invoking Bochner's theorem.

Maji and Berg (2009) presented a similar result for the intersection kernel (also known as the min kernel) $K(\mathbf{x}, \mathbf{y}) = \sum_{i=1}^{d} \min\{\mathbf{x}_i, \mathbf{y}_i\}$ which was generalized by Vedaldi and Zisserman (2010) to the class of additive homogeneous kernels $K(\mathbf{x}, \mathbf{y}) = \sum_{i=1}^{d} k_i(\mathbf{x}_i, \mathbf{y}_i)$ where $k_i(x, y) = (xy)^{\frac{\gamma}{2}} f_i(\log x - \log y)$ for some $\gamma \in \mathbb{R}$ and positive definite functions $f_i : \mathbb{R} \to \mathbb{R}$. Vempati et al. (2010) extended this idea to provide feature maps for RBF kernels of the form $K(\mathbf{x}, \mathbf{y}) = \exp\left(-\frac{1}{2\sigma^2}\chi^2(\mathbf{x}, \mathbf{y})\right)$ where $\chi^2$ is the Chi-squared distance measure.

There have been approaches that try to perform embeddings in a task dependent manner (see for example Perronnin et al., 2010). The idea of directly considering low-rank approximations to the Gram matrix has also been explored (see for example Bach and Jordan, 2005). However, the approaches considered in Rahimi and Recht (2007) and Vedaldi and Zisserman (2010) are the ones that most directly relate to this work.

### 2.1 Our Contribution

In this work we present feature maps approximating positive definite dot product kernels i.e kernels of the form $K(\mathbf{x}, \mathbf{y}) = f(\langle \mathbf{x}, \mathbf{y} \rangle)$ for some real valued function $f : \mathbb{R} \to \mathbb{R}$. More formally we present feature maps $Z : \mathbb{R}^d \to \mathbb{R}^D$ (where we refer to $R^d$ as the **input space** and $\mathbb{R}^D$ as the **embedding space**) such that for all $\mathbf{x}, \mathbf{y} \in \mathbb{R}^d, \langle Z(\mathbf{x}), Z(\mathbf{y}) \rangle \approx K(\mathbf{x}, \mathbf{y})$ with very high probability. We base our result on a characterization of real valued functions $f$ that yield such positive definite kernels. We also demonstrate how our methods can be extended to compositional kernels of the form $K_{\text{co}}(\mathbf{x}, \mathbf{y}) = K_{\text{dp}}(K(\mathbf{x}, \mathbf{y}))$ where $K_{\text{dp}}$ is some dot product kernel and $K$ is an arbitrary positive definite kernel.

The kernels covered by our approach include homogeneous polynomial kernels which are not covered by Vedaldi and Zisserman's treatment of homogeneous kernels as these are inseparable kernels which their approach cannot handle.

In the following, vectors shall be denoted in boldface. $\mathbf{x}_i$ denotes the $i^{th}$ Cartesian coordinate of a vector $\mathbf{x}$. $\mathcal{B}_p(\mathbf{0}, r)$ denotes the set $\left\{\mathbf{x} \in \mathcal{H} : \|\mathbf{x}\|_p \leq r\right\}$ for some inner product space $\mathcal{H}$ (or some finite dimensional Euclidean space $\mathbb{R}^d$). In particular, $\mathcal{B}_1(\mathbf{0}, 1)$ and $\mathcal{B}_2(\mathbf{0}, 1)$ denote set of points with less than unit 1-norm and 2-norm respectively. $\|\cdot\|$ without any subscripts denotes the 2-norm.

## 3 A Characterization of Positive Definite Dot Product Kernels

The result underlying our feature map constructions is a characterization of real valued functions on the real line that can be used to construct positive definite dot product kernels. This is a classical result in harmonic analysis due to Schoenberg (1942), that characterizes positive definite functions on the unit sphere in a Hilbert space. Our first observation, formalized below, is simply the fact that the restriction to the unit sphere is not crucial.

**Theorem 1.** *A function $f : \mathbb{R} \to \mathbb{R}$ defines a positive definite kernel $K : \mathcal{B}_2(\mathbf{0}, 1) \times \mathcal{B}_2(\mathbf{0}, 1) \to \mathbb{R}$ as $K : (\mathbf{x}, \mathbf{y}) \mapsto f(\langle \mathbf{x}, \mathbf{y} \rangle)$ iff $f$ is an analytic function admitting a Maclaurin expansion with only non-negative coefficients i.e. $f(x) = \sum_{n=0}^{\infty} a_n x^n, a_n \geq 0, n = 0, 1, 2, \ldots$. Here $\mathcal{B}_2(\mathbf{0}, 1) \subset \mathcal{H}$ for some Hilbert space $\mathcal{H}$.*

*Proof.* We first recollect Schoenberg's result in its original form

**Theorem 2** (Schoenberg (1942), Theorem 2). *A function $f : [-1, 1] \to \mathbb{R}$ constitutes a positive definite kernel $K : S_\infty \times S_\infty \to \mathbb{R}$, $K : (\mathbf{x}, \mathbf{y}) \mapsto f(\langle \mathbf{x}, \mathbf{y} \rangle)$ iff $f$ is an analytic function admitting a Maclaurin expansion with only non-negative coefficients i.e. $f(x) = \sum_{n=0}^{\infty} a_n x^n, a_n \geq 0, n = 0, 1, 2, \ldots$. Here $S_\infty = \{\mathbf{x} \in \mathcal{H} : \|\mathbf{x}\|_2 = 1\}$ for some Hilbert space $\mathcal{H}$.*

To see that the non-negativeness of the coefficients of the Maclaurin expansion is necessary just apply Theorem 2 to points on $S_\infty$. Since $\{\langle \mathbf{x}, \mathbf{y} \rangle : \mathbf{x}, \mathbf{y} \in \mathcal{B}_2(\mathbf{0}, 1)\} = \{\langle \mathbf{x}, \mathbf{y} \rangle : \mathbf{x}, \mathbf{y} \in S_\infty\}$, the result extends to the general case when the points are coming from $\mathcal{B}_2(\mathbf{0}, 1)$. To see that this suffices we make use of some well known facts regarding positive definite kernels (for example refer to Schölkopf and Smola, 2002).

**Fact 3.** *If $K_n, n \in \mathbb{N}$ are positive definite kernels defined on some common domain then the following statements are true*

1. *$c_m K_m + c_n K_n$ is also a positive definite kernel provided $c_m, c_n \geq 0$.*

2. *$K_m K_n$ is also a positive definite kernel.*

3. *If $\lim_{n \to \infty} K_n = K$ and $K$ is continuous then $K$ is also a positive definite kernel.*

Starting with the fact that the dot product kernel is positive definite on any Hilbert space $\mathcal{H}$, applying Fact 3.1 and Fact 3.2, we get that for every $n \in N$, the kernel $K_n(\mathbf{x}, \mathbf{y}) = \sum_{i=0}^{n} a_i \langle \mathbf{x}, \mathbf{y} \rangle^i$ is positive definite. An application of Fact 3.3 along with the fact that the Maclaurin series converges uniformly within its radius of convergence then proves the result. □

Actually Schoenberg (1942) shows that a function $f$ need only have a non-negative expansion in terms of Gegenbauer polynomials in order to yield a positive definite kernel over finite dimensional Euclidean spaces (a condition weaker than that of Theorem 1).

However, functions $f$ that do not have non-negative Maclaurin expansions are not very useful because they yield kernels that become indefinite after the dimensionality crosses a certain threshold. This is because a dot product kernel that is positive definite over all finite dimensional Euclidean spaces is also positive definite over Hilbert spaces (see the Section 3.1 for the simple proof).

Most dot product kernels used in practice (see Schölkopf and Smola, 2002) satisfy the stronger condition of the Maclaurin expansion having non-negative coefficients and our results readily apply to these.

We note that, as a corollary of Schoenberg's result, all dot product kernels are necessarily unbounded over non-compact domains. This is in stark contrast with translation invariant kernels that are always bounded (see Rudin, 1962, for a proof). Hence from now on we shall assume that our data is confined to some compact domain $\Omega \subset \mathbb{R}^d$. In order to study the behavior of our feature maps as this domain grows in size, we shall assume that $\Omega \subseteq \mathcal{B}_1(\mathbf{0}, R)$ for some $R > 0$.

We shall assume that the function $f$ is defined and differentiable on a closed interval $[-I, I]$. The value of $I$ shall be dictated by the value of $R$ chosen above. If $f$ is defined only on an open interval $(-\gamma, \gamma)$ around zero (as is the case when the Maclaurin series has a finite radius of convergence) then we can choose a scalar $c > \frac{I}{\gamma}$, define $g = f\left(\frac{x}{c}\right)$ and use $g$ to define a new kernel $K_g$. This has the implicit effect of scaling the data vectors in input space $\mathbb{R}^d$ down by a factor of $c$.

### 3.1 Positive definite dot product kernels over finite dimensional spaces

As noted in the main paper, the original result of Schoenberg characterizing functions that yield a positive definite dot product kernel over finite dimensional Euclidean spaces in terms of those admitting positive Gegenbauer expansions is not very useful in practice. This is because of two reasons. Firstly, as we shall show below, functions that have non-negative Gegenbauer expansions include those that yield positive definite kernels only up to a certain dimensionality i.e. these kernels are positive definite up to $\mathbb{R}^{d_0}$ for some fixed $d_0$ and indefinite on all Euclidean spaces of dimensionality $d > d_0$. Secondly, from an algorithmic perspective, the Gegenbauer expansions do not seem amenable to the type of feature construction methods described in this paper - this is because Gegenbauer polynomials themselves admit negative coefficients.

The result characterizing positive definite functions over Hilbert spaces in terms of positive Maclaurin expansions on the other hand is appealing for the very same reasons - functions satisfying this stronger con-dition are positive definite over all finite dimensional spaces and the method readily lends itself to feature construction methods.

**Lemma 4.** *A function $f : \mathbb{R} \to \mathbb{R}$ yields positive definite dot product kernels over all finite dimensional Euclidean spaces iff it yields positive definite dot product kernels over Hilbert spaces.*

*Proof.* We shall first prove this result for the special case of $\ell_2$, the Hilbert space of all square summable sequences. Schoenberg's result (Corollary 1) will then allow us to extend it to all Hilbert spaces. The *if* part follows readily from the observation that $\ell_2$ contains all finite dimensional Euclidean spaces as subspaces and the fact that any kernel that is positive definite over a set is positive definite over all its subsets as well.

For the *only if* part consider any set of $n$ points $S = \{\mathbf{x}_1, \mathbf{x}_2, \ldots, \mathbf{x}_n\} \subset \ell_2$. Clearly there exists an embedding $\Phi : S \to \mathbb{R}^n$ such that for all $i, j \in [n], \langle \Phi(\mathbf{x}_i), \Phi(\mathbf{x}_j) \rangle = \langle \mathbf{x}_i, \mathbf{x}_j \rangle$ (note that the left and the right hand sides are inner products over different spaces). Such an embedding can be constructed, for example, by taking the Cholesky decomposition of the Gram matrix given by the inner product on $\ell_2$ (the entries of the Gram matrix are finite by an application of Cauchy-Schwarz inequality).

Consider the matrix $A = [a_{ij}]$ where $a_{ij} = f(\langle \Phi(\mathbf{x}_i), \Phi(\mathbf{x}_j) \rangle)$. Since $f$ yields positive definite kernels over all finite dimensional Euclidean spaces, we have $A \succeq 0$. However, by the isometry of the embedding, we have $a_{ij} = f(\langle \mathbf{x}_i, \mathbf{x}_j \rangle)$. Hence, for any $n < \infty$, for any arbitrary $n$ points, the gram matrix given by $f(\langle \cdot, \cdot \rangle)$ is positive definite (here $\langle \cdot, \cdot \rangle$ is the dot product over $\ell_2$). Thus $f$ yields a positive definite kernel over $\ell_2$ as well.

To finish off the proof we now use Schoenberg's theorem to extend this result to all Hilbert spaces. If a dot product kernel is positive definite over all finite dimensional spaces then the above argument shows it to be positive definite over $\ell_2$. Hence, by Corollary 1, the function $f$ defining this kernel must have a non-negative Maclaurin's expansion. From here on an argument similar to the one used to prove the sufficiency part of Corollary 1 (using Fact 3) can be used to show that this kernel is positive definite over all Hilbert spaces.

On the other hand, if a dot product kernel is positive definite over Hilbert spaces, then we use its positive-definiteness over $\ell_2$, along with the argument used in showing the *if* part above, to prove that the kernel is positive definite over all finite dimensional Euclidean spaces. □

An easy application of Corollary 1 then gives us the following result :

**Corollary 5.** *A function $f : \mathbb{R} \to \mathbb{R}$ yields positive definite kernels over all finite dimensional Euclidean spaces iff it is an analytic function admitting a Maclaurin expansion with only non-negative coefficients.*

However, we note that even functions that have only positive Gegenbauer expansions (and not positive Maclaurin expansions) may admit low dimensional feature maps. This is indicated by the Johnson-Lindenstrauss Lemma (for example see Indyk and Motwani, 1998) that predicts the existence of low-distortion embeddings from arbitrary Hilbert spaces (thus, in particular from the reproducing kernel Hilbert spaces of these kernels) to finite dimensional Euclidean spaces. Interestingly, it is very tempting to view the constructions of Rahimi and Recht (2007) and Vedaldi and Zisserman (2010) (among others) as algorithmic versions of the Johnson-Lindenstrauss Lemma. The challenge in all such cases, however, is to make these constructions explicit, uniform, as well as algorithmically efficient.

### 3.2 Examples of Positive Definite Dot Product Kernels

The most well known dot product kernels are the polynomial kernels which are used in either a homogeneous form ($K(\mathbf{x}, \mathbf{y}) = \langle \mathbf{x}, \mathbf{y} \rangle^p$ for some $p \in \mathbb{N}$) or a non-homogeneous form ($K(\mathbf{x}, \mathbf{y}) = (\langle \mathbf{x}, \mathbf{y} \rangle + r)^p$ for some $p \in \mathbb{N}, r \in \mathbb{R}^+$). Lesser known examples include Vovk's real polynomial kernel ($K(\mathbf{x}, \mathbf{y}) = \frac{1 - \langle \mathbf{x}, \mathbf{y} \rangle^p}{1 - \langle \mathbf{x}, \mathbf{y} \rangle}$ for some $p \in \mathbb{N}$), Vovk's infinite polynomial kernel ($K(\mathbf{x}, \mathbf{y}) = \frac{1}{1 - \langle \mathbf{x}, \mathbf{y} \rangle}$) and the exponential dot product kernel ($K(\mathbf{x}, \mathbf{y}) = \exp\left(\frac{\langle \mathbf{x}, \mathbf{y} \rangle}{\sigma^2}\right)$ for some $\sigma \in \mathbb{R}$).

It is interesting to note that due to a result by Steinwart (2001), the last two kernels (Vovk's infinite kernel and exponential dot product kernel) are universal on any compact subset $S \subset \mathbb{R}^d$ which means that the space of all functions induced by them is dense in $C(S)$, the space of all continuous functions defined on $S$. The widely used Gaussian kernel is actually a normalized version of the exponential dot product kernel. However Vovk's kernels are seldom used in practice since they are expected to have poor generalization properties due to their flat spectrum as noted by Schölkopf and Smola (2002).

## 4 Random Feature Maps

Schoenberg's result naturally paves the way for a result of the kind presented in Rahimi and Recht (2007) in which we view the coefficients of the Maclaurin's ex-pansion as a positive measure defined on $\mathbb{N} \cup \{0\}$ and define estimators for each individual term of the expression. However, as we shall see, estimating higher order terms in our case will require more randomness. Thus, a set of coefficients $\{a_n\}$ defining a heavy tailed distribution would entail huge randomness costs in case the expansion has a large (or infinite) number of terms. For example the sequence $a_n = \frac{1}{n^2}$ has a linear rather than an exponential tail.

To address this issue we do not utilize the coefficients as measure values, rather we impose an external distribution on $\mathbb{N} \cup \{0\}$ having an exponential tail. The distribution that we choose to impose is $\mathbb{P}[N = n] = \frac{1}{p^{n+1}}$ for some fixed $p > 1$. In practice $p = 2$ is a good choice since it establishes a normalized measure over $\mathbb{N} \cup \{0\}$. We will, using this distribution, obtain unbiased estimates for the kernel value and prove corresponding uniform convergence results.

We stress that the positiveness of the coefficients $\{a_n\}$ is still essential for us to be able to provide an embedding into real spaces. If the coefficients are allowed to be negative, the resulting kernels would no longer remain positive definite and we would only be able to provide feature maps that map to pseudo-Euclidean spaces. It turns out that the imposition of an external measure is crucial from a statistical point of view as well. As we shall see later, it allows us to obtain bounded estimators which in turn allow us to use Hoeffding bounds to prove uniform convergence results.

We now move on to describe our feature map : our feature map will essentially be a concatenation of several copies of identical real valued feature maps. These copies will reduce variance and allow us to prove convergence bounds. The following simple fact about random projections is at the core of our feature maps.

**Lemma 6.** *Let $\boldsymbol{\omega} \in \mathbb{R}^d$ be a vector each of whose coordinates have been chosen pairwise independently using fair coin tosses from the set $\{-1, 1\}$ and consider the feature map $Z : \mathbb{R}^d \to \mathbb{R}, Z : \mathbf{x} \mapsto \boldsymbol{\omega}^\top \mathbf{x}$. Then for all $\mathbf{x}, \mathbf{y} \in \mathbb{R}^d, \mathbb{E}_{\boldsymbol{\omega}}[Z(\mathbf{x})Z(\mathbf{y})] = \langle \mathbf{x}, \mathbf{y} \rangle$.*

*Proof.* We have $\mathbb{E}_{\boldsymbol{\omega}}[Z(\mathbf{x})Z(\mathbf{y})] = \mathbb{E}_{\boldsymbol{\omega}}\left[\boldsymbol{\omega}^\top \mathbf{x} \cdot \boldsymbol{\omega}^\top \mathbf{y}\right]$

$$= \mathbb{E}_{\boldsymbol{\omega}}\left[\left(\sum_{i=1}^d \boldsymbol{\omega}_i \mathbf{x}_i\right)\left(\sum_{i=1}^d \boldsymbol{\omega}_i \mathbf{y}_i\right)\right]$$

$$= \mathbb{E}_{\boldsymbol{\omega}}\left[\sum_{i=1}^d \boldsymbol{\omega}_i^2 \mathbf{x}_i \mathbf{y}_i + \sum_{i \neq j}^d \boldsymbol{\omega}_i \boldsymbol{\omega}_j \mathbf{x}_i \mathbf{y}_j\right]$$

$$= \sum_{i=1}^d \mathbb{E}_{\boldsymbol{\omega}}\left[\boldsymbol{\omega}_i^2\right] \mathbf{x}_i \mathbf{y}_i + \sum_{i \neq j}^d \mathbb{E}_{\boldsymbol{\omega}}[\boldsymbol{\omega}_i]\mathbb{E}_{\boldsymbol{\omega}}[\boldsymbol{\omega}_j] \mathbf{x}_i \mathbf{y}_j$$

$$= \sum_{i=1}^d \mathbf{x}_i \mathbf{y}_i + 0 = \langle \mathbf{x}, \mathbf{y} \rangle$$

where in the third equality we have used linearity of expectation and the pairwise independence of the different coordinates of $\boldsymbol{\omega}$. The fourth equality is arrived at by using properties of the distribution. Notice that any distribution that is symmetric about zero with unit second moment can be used for sampling the coordinates of $\boldsymbol{\omega}$. This particular choice both simplifies the analysis as well as is easy to implement in practice. □

We now present a real valued feature map for the dot product kernel. First of all we randomly pick a number $N \in \mathbb{N} \cup \{0\}$ with $\mathbb{P}[N = n] = \frac{1}{p^{n+1}}$. Next we pick $N$ independent Rademacher vectors $\boldsymbol{\omega}_1 \ldots \boldsymbol{\omega}_N$ and output the feature map $Z : \mathbb{R}^d \to \mathbb{R}$, $Z : \mathbf{x} \mapsto \sqrt{a_N p^{N+1}} \prod_{j=1}^N \boldsymbol{\omega}_j^\top \mathbf{x}$. We first of all establish that the linear kernel obtained by using this feature map gives us an unbiased estimate of the kernel value at each pair of points chosen from the domain $\Omega$.

**Lemma 7.** *Let $Z : \mathbb{R}^d \to \mathbb{R}$ be the feature map constructed above. Then for all $\mathbf{x}, \mathbf{y} \in \Omega$, we have $\mathbb{E}[Z(\mathbf{x})Z(\mathbf{y})] = K(\mathbf{x}, \mathbf{y})$ where the expectation is over the choice of the Rademacher vectors.*

*Proof.* We have $\mathbb{E}[Z(\mathbf{x})Z(\mathbf{y})]$

$$= \mathbb{E}_N\left[\mathbb{E}_{\boldsymbol{\omega}_1,\ldots,\boldsymbol{\omega}_N}[Z(\mathbf{x})Z(\mathbf{y})]\Big| N\right]$$

$$= \mathbb{E}_N\left[a_N p^{N+1} \mathbb{E}_{\boldsymbol{\omega}_1,\ldots,\boldsymbol{\omega}_N}\left[\prod_{j=1}^N \boldsymbol{\omega}_j^\top \mathbf{x} \prod_{j=1}^N \boldsymbol{\omega}_j^\top \mathbf{y}\right]\right]$$

$$= \mathbb{E}_N\left[a_N p^{N+1} \left(\mathbb{E}_{\boldsymbol{\omega}}\left[\boldsymbol{\omega}^\top \mathbf{x} \cdot \boldsymbol{\omega}^\top \mathbf{y}\right]\right)^N\right]$$

$$= \mathbb{E}_N\left[a_N p^{N+1} \langle \mathbf{x}, \mathbf{y} \rangle^N\right]$$

$$= \sum_{n=0}^\infty \frac{1}{p^{n+1}} \cdot a_n p^{n+1} \langle \mathbf{x}, \mathbf{y} \rangle^n$$

$$= K(\mathbf{x}, \mathbf{y}).$$

where the first step uses the fact that the index $N$ and the vectors $\boldsymbol{\omega}_i$ are chosen independently, the fourth step uses the fact that the vectors $\boldsymbol{\omega}_i$ are chosen independently among themselves and the fifth step uses Lemma 2. □

Having obtained a feature map giving us an unbiased estimate of the kernel value, we move on to establish bounds on the deviation of the linear kernel given by this map from its expected value. To do this we obtain $D$ such feature maps independently and concatenate them to obtain a multi dimensional feature map $\mathbf{Z} : \mathbb{R}^d \to \mathbb{R}^D, \mathbf{Z} : \mathbf{x} \mapsto \frac{1}{\sqrt{D}}(Z_1(\mathbf{x}), \ldots, Z_D(\mathbf{x}))$. It is easy to see that $\mathbb{E}\left[\langle \mathbf{Z}(\mathbf{x}), \mathbf{Z}(\mathbf{y})\rangle\right] = K(\mathbf{x}, \mathbf{y})$. Moreover, such a concatenation is expected to guarantee an exponentially fast convergence to $K(\mathbf{x}, \mathbf{y})$ using Hoeffding bounds. However this requires us to prove that the estimator corresponding to our feature map i.e $Z(\mathbf{x})Z(\mathbf{y})$ is bounded. This we establish below :

**Lemma 8.** *For all* $\mathbf{x}, \mathbf{y} \in \Omega, |Z(\mathbf{x})Z(\mathbf{y})| \leq pf(pR^2)$.

*Proof.* Since $Z(\mathbf{x})Z(\mathbf{y}) = a_N p^{N+1} \prod_{j=1}^{N} \boldsymbol{\omega}_j^\top \mathbf{x} \prod_{j=1}^{N} \boldsymbol{\omega}_j^\top \mathbf{y}$, by Hölder's inequality we have, for all $j, |\boldsymbol{\omega}_j^\top \mathbf{x}| \leq \|\boldsymbol{\omega}_j\|_\infty \|\mathbf{x}\|_1 \leq R$ since every coordinate of $\boldsymbol{\omega}_j$ is either 1 or $-1$ and $\mathbf{x} \in \Omega \subseteq \mathcal{B}_1(\mathbf{0}, R)$. A similar result holds for $|\boldsymbol{\omega}_j^\top \mathbf{y}|$ as well. Thus we have $|Z(\mathbf{x})Z(\mathbf{y})| \leq a_N p^{N+1} R^{2N} \leq p \cdot \sum_{n=0}^{\infty} a_n p^n R^{2n} = pf(pR^2)$. □

We note here that the imposition of an external measure on $\mathbb{N} \cup \{0\}$ plays a crucial role in the analysis. In absence of the external measure, one is only able to bound the estimator by $\emptyset R^{2N}$ and since $N$ is a potentially unbounded random variable, this makes application of Hoeffding bounds impossible. Although there do exist Hoeffding style bounds for unbounded random variables, none seem to work in our case. However, with the simple imposition of an external measure we obtain an estimator that is bounded by a value dependent on the range of values taken by the kernel over the domain, a very desirable quality.

For sake of convenience let us denote $pf(pR^2)$ by $C_\Omega$ since it is a constant dependent only on the size of the domain $\Omega$ and independent of the dimension of the input space $\mathbb{R}^d$. Note that this constant is proportional to the largest value taken by the kernel in the domain $\Omega$. This immediately tells us that for any $\mathbf{x}, \mathbf{y} \in \Omega$, $\mathbb{P}\left[|\langle \mathbf{Z}(\mathbf{x}), \mathbf{Z}(\mathbf{y})\rangle - K(\mathbf{x}, \mathbf{y})| > \epsilon\right] \leq 2\exp\left(-\frac{D\epsilon^2}{8C_\Omega^2}\right)$. However we can give much stronger guarantees than this – we can prove that this loss of confidence need not be incurred over every single pair of points but rather the entire domain at once. More formally, we can show that with very high probability, $\sup_{\mathbf{x}, \mathbf{y} \in \Omega} |\langle \mathbf{Z}(\mathbf{x}), \mathbf{Z}(\mathbf{y})\rangle - K(\mathbf{x}, \mathbf{y})| \leq \epsilon$.

### 4.1 Uniform Approximation

As stated before, we are able to ensure that the feature map designed above gives an accurate estimate of the kernel value uniformly over the entire domain. For this we exploit the Lipschitz properties of the ker-

**Algorithm 1** Random Maclaurin Feature Maps

**Require:** A positive definite dot product kernel $K(\mathbf{x}, \mathbf{y}) = f(\langle \mathbf{x}, \mathbf{y}\rangle)$.
**Ensure:** A randomized feature map $\mathbf{Z} : \mathbb{R}^d \to \mathbb{R}^D$ such that $\langle \mathbf{Z}(\mathbf{x}), \mathbf{Z}(\mathbf{y})\rangle \approx K(\mathbf{x}, \mathbf{y})$.

Obtain the Maclaurin expansion of $f(x) = \sum_{n=0}^{\infty} a_n x^n$ by setting $a_n = \frac{f^{(n)}(0)}{n!}$.
Fix a value $p > 1$.
**for** $i = 1$ **to** $D$ **do**
  Choose a non negative integer $N \in \mathbb{N} \cup \{0\}$ with $\mathbb{P}[N = n] = \frac{1}{p^{n+1}}$.
  Choose $N$ vectors $\boldsymbol{\omega}_1, \ldots, \boldsymbol{\omega}_N \in \{-1, 1\}^d$ selecting each coordinate using fair coin tosses.
  Let feature map $Z_i : \mathbf{x} \mapsto \sqrt{a_N p^{N+1}} \prod_{j=1}^{N} \boldsymbol{\omega}_j^\top \mathbf{x}$.
**end for**
Output $\mathbf{Z} : \mathbf{x} \mapsto \frac{1}{\sqrt{D}}(Z_1(\mathbf{x}), \ldots, Z_D(\mathbf{x}))$.

---

nel function and our estimator. A similar approach was adopted by Rahimi and Recht (2007) to provide corresponding uniform convergence properties for their estimator. However it is not possible to import their argument since they were able to exploit the fact that both their kernel as well as their estimator were translation invariant. We, having no such guarantees for our estimator, have to argue differently.

Let $\mathcal{E}(\mathbf{x}, \mathbf{y}) = \langle \mathbf{Z}(\mathbf{x}), \mathbf{Z}(\mathbf{y})\rangle - K(\mathbf{x}, \mathbf{y})$. We will first show that the function $\mathcal{E}(\cdot, \cdot)$ is Lipschitz over the domain $\Omega$. Since $\mathcal{E}(\cdot, \cdot)$ itself is differentiable (actually analytic), its Lipschitz constant can be bounded by bounding the norms of its gradients i.e. it would suffice to show that $\sup_{\mathbf{x}, \mathbf{y} \in \Omega} \|\nabla_\mathbf{x} \mathcal{E}(\mathbf{x}, \mathbf{y})\| \leq L$ and $\sup_{\mathbf{x}, \mathbf{y} \in \Omega} \|\nabla_\mathbf{y} \mathcal{E}(\mathbf{x}, \mathbf{y})\| \leq L$ for some constant $L$. This would ensure that if the error incurred by the feature map is small on a pair of vectors then it would also be small on all pairs of vectors that are "close" to these vectors. This is formalized in the following theorem :

**Lemma 9.** *If a bivariate function $f$ defined over $\Omega \subseteq \mathbb{R}^d$ is $L$-Lipschitz in both its arguments then for every $\mathbf{x}, \mathbf{y} \in \Omega$,* $\sup_{\substack{\mathbf{x}' \in \mathcal{B}_2(\mathbf{x}, r) \cap \Omega \\ \mathbf{y}' \in \mathcal{B}_2(\mathbf{y}, r) \cap \Omega}} |f(\mathbf{x}, \mathbf{y}) - f(\mathbf{x}', \mathbf{y}')| \leq 2Lr$.

*Proof.* We have $|f(\mathbf{x}, \mathbf{y}) - f(\mathbf{x}', \mathbf{y}')| \leq |f(\mathbf{x}, \mathbf{y}) - f(\mathbf{x}, \mathbf{y}')| + |f(\mathbf{x}, \mathbf{y}') - f(\mathbf{x}', \mathbf{y}')| \leq L \cdot \|\mathbf{y} - \mathbf{y}'\| + L \cdot \|\mathbf{x} - \mathbf{x}'\| \leq 2Lr$ where in the second step we have used the fact that $\mathbf{x}, \mathbf{y}' \in \Omega$. □

What this allows us to do is choose a set of points $\mathcal{T}$ that set up an $\epsilon$-net over the domain $\Omega$ at some scale $\epsilon_1$. If we can ensure that the feature maps pro-vide an ($\epsilon/2$)-close approximation to $K$ at the centers of this net i.e. $\sup_{\mathbf{x},\mathbf{y}\in\mathcal{T}} |\mathcal{E}(\mathbf{x},\mathbf{y})| \leq \epsilon/2$, then the above result would show us that if the error function $\mathcal{E}(\cdot,\cdot)$ is $L$-Lipschitz in both its arguments, then $\sup_{\mathbf{x},\mathbf{y}\in\Omega} |\mathcal{E}(\mathbf{x},\mathbf{y})| \leq \epsilon/2 + 2L\epsilon_1$ since the $\epsilon$-net ensures that for all $\mathbf{x},\mathbf{y}\in\Omega$, there exists $\mathbf{x}',\mathbf{y}'\in\mathcal{T}$ such that $\|\mathbf{x}-\mathbf{x}'\|, \|\mathbf{y}-\mathbf{y}'\| \leq \epsilon_1$. Thus choosing $\epsilon_1 = \frac{\epsilon}{4L}$ ensures that $\sup_{\mathbf{x},\mathbf{y}\in\Omega} |\langle \mathbf{Z}(\mathbf{x}), \mathbf{Z}(\mathbf{y})\rangle - K(\mathbf{x},\mathbf{y})| \leq \epsilon$.

Now ensuring that the feature maps provide a close approximation to the kernel value at all pairs of points taken from $\mathcal{T}$ would cost us a reduction in the confidence parameter by a factor of $|\mathcal{T}|^2$ due to taking a union bound. It is well known (for example see Cucker and Smale, 2001) that setting up an $\epsilon$-net at scale $\epsilon_1$ in $d$ dimensions over a compact set of diameter $\Delta$ takes at most $\left(\frac{4\Delta}{\epsilon_1}\right)^d$ centers. In our case $\Delta \leq 2R$ since $\Omega \subseteq \mathcal{B}_1(\mathbf{0},R) \subset \mathcal{B}_2(\mathbf{0},R)$ and $\epsilon_1 = \frac{\epsilon}{4L}$ i.e. $|\mathcal{T}| \leq \left(\frac{32RL}{\epsilon}\right)^d$.

We now move on to the task of bounding the Lipschitz constant of the error function. Since $\mathcal{E}(\cdot,\cdot)$ is symmetric in both its arguments, it is sufficient to bound $\|\nabla_\mathbf{x}\mathcal{E}(\mathbf{x},\mathbf{y})\| \leq \|\nabla_\mathbf{x}\langle\mathbf{Z}(\mathbf{x}),\mathbf{Z}(\mathbf{y})\rangle\| + \|\nabla_\mathbf{x} K(\mathbf{x},\mathbf{y})\|$. We will bound these two quantities separately below.

**Lemma 10.** *We have the following :*
$$\sup_{\mathbf{x},\mathbf{y}\in\Omega} \|\nabla_\mathbf{x} K(\mathbf{x},\mathbf{y})\| \leq Rf'(R^2)$$
$$\sup_{\mathbf{x},\mathbf{y}\in\Omega} \|\nabla_\mathbf{y} K(\mathbf{x},\mathbf{y})\| \leq Rf'(R^2)$$

*Proof.* We have $\nabla_\mathbf{x} K(\mathbf{x},\mathbf{y}) = \nabla_\mathbf{x}\left(\sum_{n=0}^{\infty} a_n \langle\mathbf{x},\mathbf{y}\rangle^n\right) = \sum_{n=0}^{\infty} a_n \nabla_\mathbf{x}\langle\mathbf{x},\mathbf{y}\rangle^n = \mathbf{y} \sum_{n=0}^{\infty} na_n \langle\mathbf{x},\mathbf{y}\rangle^{n-1}$. Thus we have $\|\nabla_\mathbf{x} K(\mathbf{x},\mathbf{y})\| = \left\|\mathbf{y}\sum_{n=0}^{\infty} na_n \langle\mathbf{x},\mathbf{y}\rangle^{n-1}\right\| \leq R\sum_{n=0}^{\infty} na_n |\langle\mathbf{x},\mathbf{y}\rangle|^{n-1} \leq R\sum_{n=0}^{\infty} na_n (R^2)^{n-1} = Rf'(R^2)$ where in the second and the third step we have used the fact that $\mathbf{x},\mathbf{y}\in\Omega \subseteq \mathcal{B}_1(\mathbf{0},R) \subset \mathcal{B}_2(\mathbf{0},R)$. Similarly we can show $\sup_{\mathbf{x},\mathbf{y}\in\Omega} \|\nabla_\mathbf{y} K(\mathbf{x},\mathbf{y})\| \leq Rf'(R^2)$. □

**Lemma 11.** *We have the following :*
$$\sup_{\mathbf{x},\mathbf{y}\in\Omega} \|\nabla_\mathbf{x} (Z_1(\mathbf{x})Z_1(\mathbf{y}))\| \leq p^2 R\sqrt{d} f'(pR^2)$$
$$\sup_{\mathbf{x},\mathbf{y}\in\Omega} \|\nabla_\mathbf{y} (Z_1(\mathbf{x})Z_1(\mathbf{y}))\| \leq p^2 R\sqrt{d} f'(pR^2)$$

*Proof.* Since $\langle\mathbf{Z}(\mathbf{x}),\mathbf{Z}(\mathbf{y})\rangle = \frac{1}{D}\sum_{i=1}^{D} Z_i(\mathbf{x})Z_i(\mathbf{y})$ and $\nabla_\mathbf{x}\langle\mathbf{Z}(\mathbf{x}),\mathbf{Z}(\mathbf{y})\rangle = \frac{1}{D}\sum_{i=1}^{D} \nabla_\mathbf{x}(Z_i(\mathbf{x})Z_i(\mathbf{y}))$ we have $\|\nabla_\mathbf{x}\langle\mathbf{Z}(\mathbf{x}),\mathbf{Z}(\mathbf{y})\rangle\| \leq \frac{1}{D}\sum_{i=1}^{D} \|\nabla_\mathbf{x}(Z_i(\mathbf{x})Z_i(\mathbf{y}))\|$ by triangle inequality. Since all the $Z_i$ feature maps are identical it would be sufficient to bound $\|\nabla_\mathbf{x}(Z_1(\mathbf{x})Z_1(\mathbf{y}))\|$ and by the above calculation, the same bound would hold for $\|\nabla_\mathbf{x}\langle\mathbf{Z}(\mathbf{x}),\mathbf{Z}(\mathbf{y})\rangle\|$ as well. Let $Z_1 : \mathbf{x} \mapsto \sqrt{a_N p^{N+1}} \prod_{j=1}^{N} \boldsymbol{\omega}_j^\top \mathbf{x}$ for some $N \leq k$.

Thus we can bound the quantity $\nabla_\mathbf{x}(Z_1(\mathbf{x})Z_1(\mathbf{y}))$ as $\nabla_\mathbf{x}\left(a_N p^{N+1}\prod_{j=1}^{N}\boldsymbol{\omega}_j^\top\mathbf{x}\prod_{j=1}^{N}\boldsymbol{\omega}_j^\top\mathbf{y}\right)$ which simplifies to $\left(a_N p^{N+1}\prod_{j=1}^{N}\boldsymbol{\omega}_j^\top\mathbf{y}\right)\nabla_\mathbf{x}\left(\prod_{j=1}^{N}\boldsymbol{\omega}_j^\top\mathbf{x}\right)$ and further to $\left(a_N p^{N+1}\prod_{j=1}^{N}\boldsymbol{\omega}_j^\top\mathbf{y}\right)\sum_{j=1}^{N}\left(\prod_{i\neq j}\boldsymbol{\omega}_i^\top\mathbf{x}\right)\boldsymbol{\omega}_i$.

We note that for any $\boldsymbol{\omega}$ chosen, $\|\boldsymbol{\omega}\| = \sqrt{d}$. Moreover, as we have seen before, for any $\boldsymbol{\omega}, \sup_{\mathbf{x}\in\Omega} |\boldsymbol{\omega}^\top\mathbf{x}| \leq R$ by Hölder's inequality. Thus we can bound $\|\nabla_\mathbf{x}(Z_1(\mathbf{x})Z_1(\mathbf{y}))\|$ as

$$\left\|\left(a_N p^{N+1}\prod_{j=1}^{N}\boldsymbol{\omega}_j^\top\mathbf{y}\right)\sum_{j=1}^{N}\left(\prod_{i\neq j}\boldsymbol{\omega}_i^\top\mathbf{x}\right)\boldsymbol{\omega}_i\right\|$$
$$= a_N p^{N+1}\left(\prod_{j=1}^{N}|\boldsymbol{\omega}_j^\top\mathbf{y}|\right)\left\|\sum_{j=1}^{N}\left(\prod_{i\neq j}\boldsymbol{\omega}_i^\top\mathbf{x}\right)\boldsymbol{\omega}_i\right\|$$
$$\leq a_N p^{N+1}\left(\prod_{j=1}^{N}|\boldsymbol{\omega}_j^\top\mathbf{y}|\right)\sum_{j=1}^{N}\left(\prod_{i\neq j}|\boldsymbol{\omega}_i^\top\mathbf{x}|\right)\|\boldsymbol{\omega}_i\|$$
$$\leq a_N p^{N+1} R^N \sum_{j=1}^{N} R^{N-1}\sqrt{d} = Na_N p^{N+1} R^{2N-1}\sqrt{d}$$
$$\leq p^2 R\sqrt{d}\sum_{n=0}^{\infty} na_n(pR^2)^{n-1} = p^2 R\sqrt{d} f'(pR^2)$$

where we have used the triangle inequality in the third step. Similarly we can show $\sup_{\mathbf{x},\mathbf{y}\in\Omega}\|\nabla_\mathbf{y}(Z_1(\mathbf{x})Z_1(\mathbf{y}))\| \leq p^2 R\sqrt{d} f'(pR^2)$. □

Thus we have $L = \sup_{\mathbf{x},\mathbf{y}\in\Omega}\|\nabla_\mathbf{x}\mathcal{E}(\mathbf{x},\mathbf{y})\| \leq Rf'(R^2) + p^2 R\sqrt{d} f'(pR^2)$. Putting all the results together, we first have by application of union bound that the probability that the feature map will fail at any pair of points chosen from the $\epsilon$-net is bounded by $2\left(\frac{32RL}{\epsilon}\right)^{2d}\exp\left(-\frac{D\epsilon^2}{8C_\Omega^2}\right)$. The covering argument along with the bound on the Lipschitz constant of the error function ensure that with the same confidence, the feature map would provide an $\epsilon$-accurate estimate on the entire domain $\Omega$. Thus we have the following theorem.

Random Feature Maps for Dot Product Kernels

**Theorem 12.** *Let $\Omega \subseteq \mathcal{B}_1(\mathbf{0}, R)$ be a compact subset of $\mathbb{R}^d$ and $K(\mathbf{x}, \mathbf{y}) = f(\langle \mathbf{x}, \mathbf{y} \rangle)$ be a dot product kernel defined on $\Omega$. Then, for the feature map $\mathbf{Z}$ defined in Algorithm 1, we have $\mathbb{P}\left[ \sup_{\mathbf{x}, \mathbf{y} \in \Omega} |\langle \mathbf{Z}(\mathbf{x}), \mathbf{Z}(\mathbf{y}) \rangle - K(\mathbf{x}, \mathbf{y})| > \epsilon \right] \leq 2 \left(\frac{32RL}{\epsilon}\right)^{2d} \exp\left(-\frac{D\epsilon^2}{8C_\Omega^2}\right)$ where $C_\Omega = pf(pR^2)$ and $L = Rf'(R^2) + p^2R\sqrt{d}f'(pR^2)$ for some small constant $p > 1$. Moreover, with $D = \Omega\left(\frac{dC_\Omega^2}{\epsilon^2} \log\left(\frac{RL}{\epsilon\delta}\right)\right)$, one can ensure the same with probability greater than $1 - \delta$.*

The behavior of this bound with respect to the dimensionality of the input space, the accuracy parameter and the confidence parameter is of the form $D = \Omega\left(\frac{d}{\epsilon^2} \log\left(\frac{1}{\epsilon\delta}\right)\right)$ that matches that of Rahimi and Recht (2007). The bound has a stronger dependence on kernel specific parameters which appear as non-logarithmic terms due to the unbounded nature of the dot product kernels. Even so, the kernel specific term $C_\Omega$ is dependent on the largest value taken by the kernel in the domain $\Omega$, a dependence that is unavoidable for an algorithm giving guarantees on the absolute (rather than relative) deviation from the true value.

### 4.2 An Alternative Feature Map

An alternative method to bounding the amount of randomness being used is to truncate the Maclaurin series after a certain number of terms and use the resulting function to define a new kernel. Since the Maclaurin series of an analytic function defined over a bounded domain converges to it uniformly, we can truncate the series while incurring a uniformly bounded error. A similar approach is used in Vedaldi and Zisserman (2010) to present deterministic feature maps. Suppose we have a positive definite dot product kernel $K$ defined on a domain $\Omega \subset \mathcal{B}_1(\mathbf{0}, R)$ in some Euclidean space $\mathbb{R}^d$ by a function $f(x) = \sum_{n=0}^{\infty} a_n x^n$. If we choose $k = k(\epsilon, R)$ such that $\sum_{n=0}^{k} a_n R^{2n} = f(R^2) - \epsilon$ (or select some set $S \subset \mathbb{N} \cup \{0\}$ such that $\sum_{n \in S} a_n R^{2n} = f(R^2) - \epsilon$ and $|S| = k$) and create a new kernel $\tilde{K}(\mathbf{x}, \mathbf{y}) = \sum_{n=0}^{k} a_n \langle \mathbf{x}, \mathbf{y} \rangle^n$, then the residual error $R_k = \sup_{\mathbf{x}, \mathbf{y} \in \Omega} \left| \tilde{K}(\mathbf{x}, \mathbf{y}) - K(\mathbf{x}, \mathbf{y}) \right| = \sup_{\mathbf{x}, \mathbf{y} \in \Omega} \left| \sum_{i=k+1}^{\infty} a_n \langle \mathbf{x}, \mathbf{y} \rangle^n \right| \leq \sum_{i=k+1}^{\infty} a_n R^{2n} \leq \epsilon$ since $\Omega \subset \mathcal{B}_1(\mathbf{0}, R) \subset \mathcal{B}_2(\mathbf{0}, R)$ and $\sum_{n=0}^{\infty} a_n R^{2n} = f(R^2)$. Thus for all $\mathbf{x}, \mathbf{y} \in \Omega$, $K(\mathbf{x}, \mathbf{y}) - \epsilon \leq \tilde{K}(\mathbf{x}, \mathbf{y}) \leq K(\mathbf{x}, \mathbf{y}) + \epsilon$. Since $\tilde{K}$ also satisfies the conditions of Corollary 1, one can now obtain $\epsilon_1$-accurate feature maps for $\tilde{K}$ using the techniques mentioned above and those feature maps would provide an $(\epsilon + \epsilon_1)$-accurate estimate to $K$.

## 5 Generalizing to Compositional Kernels

Given a positive definite dot product kernel $K_{\text{dp}}$ and an arbitrary positive definite kernel $K$, the kernel $K_{\text{co}}$ defined as $K_{\text{co}}(\mathbf{x}, \mathbf{y}) = K_{\text{dp}}(K(\mathbf{x}, \mathbf{y}))$ is also positive definite. This fact can be deduced either by directly invoking a result due to FitzGerald et al. (1995, Theorem 2.1) or by applying Schoenberg's result in conjunction with Mercer's theorem. We now show how to extend the result for dot product kernels to such compositional kernels.

Note that plugging a translation invariant kernel into a dot product kernel yields yet another translation invariant kernel since the set of translation invariant kernels is closed under powering, scalar multiplication and addition. However, a set of homogeneous kernels not sharing the homogeneity parameter is not closed under addition. Hence the set of homogeneous kernels is not closed under the operations mentioned above and thus, plugging a homogeneous kernel into a dot product kernel in general yields a novel non-homogeneous kernel. We also note that the results obtained in the section above can be now viewed as special cases of the result presented in this section with the dot product being substituted into a dot product kernel.

In order to construct feature maps for the compositional kernel we assume that we have black-box access to a (possibly randomized) feature map selection routine $\mathcal{A}$ which when invoked, returns a feature map $W : \mathbb{R}^d \to \mathbb{R}$ for $K$. If we assume that the kernel $K$ is bounded and Lipschitz and that the feature map $W$ returned to us is bounded, Lipschitz on expectation and provides an unbiased estimate of $K$, then one can design (using these feature maps for $K$) feature maps for $K_{\text{co}}$. The analysis of the final feature map in this case is a bit more involved since we only assume black-box access to $\mathcal{A}$ and only expect the feature map to be Lipschitz on expectation.

We first formally state the assumptions made about the kernel $K$ and the feature maps returned by $\mathcal{A}$:

1. $K$ is defined over some domain $\Omega \subset \mathbb{R}^d$.

2. $K$ is bounded i.e. we have $\sup_{\mathbf{x}, \mathbf{y} \in \Omega} |K(\mathbf{x}, \mathbf{y})| \leq C_K$ for some $C_K \in \mathbb{R}^+$.

3. $K$ is Lipschitz i.e. we have $\sup_{\mathbf{x}, \mathbf{y} \in \Omega} \|\nabla_\mathbf{x} K(\mathbf{x}, \mathbf{y})\| \leq$

Purushottam Kar, Harish Karnick**Algorithm 2** Random Maclaurin Feature Maps for Compositional Kernels

**Require:** A compositional positive definite kernel $K_{\text{co}}(\mathbf{x}, \mathbf{y}) = K_{\text{dp}}(K(\mathbf{x}, \mathbf{y})) = f(K(\mathbf{x}, \mathbf{y}))$.
**Ensure:** A randomized feature map $\mathbf{Z} : \mathbb{R}^d \to \mathbb{R}^D$ such that $\langle \mathbf{Z}(\mathbf{x}), \mathbf{Z}(\mathbf{y}) \rangle \approx K_{\text{co}}(\mathbf{x}, \mathbf{y})$.

Obtain the Maclaurin expansion of $f(x) = \sum_{n=0}^{\infty} a_n x^n$ by setting $a_n = \frac{f^{(n)}(0)}{n!}$.
Fix a value $p > 1$.
**for** $i = 1$ **to** $D$ **do**
    Choose a non negative integer $N \in \mathbb{N} \cup \{0\}$ with $\mathbb{P}[N = n] = \frac{1}{p^{n+1}}$.
    Get $N$ independent instantiations of the feature map for $K$ from $\mathcal{A}$ as $W_1, \ldots, W_N$.
    Let feature map $Z_i : \mathbf{x} \mapsto \sqrt{a_N p^{N+1}} \prod_{j=1}^{N} W_j(\mathbf{x})$.
**end for**
Output $\mathbf{Z} : \mathbf{x} \mapsto \frac{1}{\sqrt{D}} (Z_1(\mathbf{x}), \ldots, Z_D(\mathbf{x}))$.

$L_K$ and $\sup_{\mathbf{x},\mathbf{y}\in\Omega} \|\nabla_{\mathbf{y}} K(\mathbf{x}, \mathbf{y})\| \leq L_K$ for some $L_K \in \mathbb{R}^+$.

4. $W$ is an unbiased estimator of $K$ i.e. for all $\mathbf{x}, \mathbf{y} \in \Omega$, $\mathbb{E}[W(\mathbf{x})W(\mathbf{y})] = K(\mathbf{x}, \mathbf{y})$ where the expectation is over the internal randomness of $W$.

5. $W$ is a bounded feature map i.e. there exists some $C_W \in \mathbb{R}^+$ such that $\sup_{\mathbf{x}\in\Omega} |W(\mathbf{x})| \leq \sqrt{C_W}$.

6. $W$ is Lipschitz on expectation i.e. for some $L_W \in \mathbb{R}^+$, $\sup_{\mathbf{x}\in\Omega} \mathbb{E}[\|\nabla_{\mathbf{x}} W(\mathbf{x})\|] \leq L_W$.

Our feature map construction algorithm is similar to the one used for dot product kernels. We pick a non-negative integer $N \in \mathbb{N} \cup \{0\}$ with $\mathbb{P}[N = n] = \frac{1}{p^{n+1}}$ for some fixed $p > 1$ and output the feature map $Z : \mathbb{R}^d \to \mathbb{R}$, $Z : \mathbf{x} \mapsto \sqrt{a_N p^{N+1}} \prod_{j=1}^{N} W_j(\mathbf{x})$ where $W_1, \ldots, W_N$ are independent instantiations of the feature map $W$ associated with the kernel $K$. We concatenate $D$ such feature maps to give our final feature map.

It is clear that on expectation, the product of the feature map values is equal to the value of the kernel i.e. $\mathbb{E}_{N,W_1,\ldots,W_N} [\langle \mathbf{Z}(\mathbf{x}), \mathbf{Z}(\mathbf{y}) \rangle] = K_{comp}(\mathbf{x}, \mathbf{y})$ where $\mathbf{Z} : \mathbb{R}^d \to \mathbb{R}^D$, $\mathbf{Z} : \mathbf{x} \mapsto \frac{1}{\sqrt{D}} (Z_1(\mathbf{x}), \ldots, Z_D(\mathbf{x}))$. Yet again we expect that the concatenation of $D$ such feature maps for a large enough $D$ would provide us a close approximation to $K_{\text{co}}$ with high probability. For this we first prove that our feature map is bounded.

**Lemma 13.** *For all $\mathbf{x}, \mathbf{y} \in \Omega$, $|Z(\mathbf{x})Z(\mathbf{y})| \leq pf(pC_W)$.*

*Proof.* $Z(\mathbf{x})Z(\mathbf{y}) = a_N p^{N+1} \prod_{j=1}^{N} W_j(\mathbf{x}) \prod_{j=1}^{N} W_j(\mathbf{x})$. Using the bound on the feature maps we get the inequality $|Z(\mathbf{x})Z(\mathbf{y})| \leq a_N p^{N+1} C_W^N \leq pf(pC_W)$. □

Thus we have for any $\mathbf{x}, \mathbf{y} \in \Omega$, $\mathbb{P}[|\langle \mathbf{Z}(\mathbf{x}), \mathbf{Z}(\mathbf{y}) \rangle - K_{\text{co}}(\mathbf{x}, \mathbf{y})| \leq \epsilon]$ with probability at least $1 - 2\exp\left(-\frac{D\epsilon^2}{8C_1^2}\right)$ where $C_1 = pf(pC_W)$. We now investigate the Lipschitz properties of $K_{\text{co}}$ and our feature map.

**Lemma 14.** *We have*

$$\sup_{\mathbf{x},\mathbf{y}\in\Omega} \|\nabla_{\mathbf{x}} K_{co}(\mathbf{x}, \mathbf{y})\| \leq L_K f'(C_K)$$
$$\sup_{\mathbf{x},\mathbf{y}\in\Omega} \|\nabla_{\mathbf{y}} K_{co}(\mathbf{x}, \mathbf{y})\| \leq L_K f'(C_K)$$

*Proof.* $K_{\text{comp}}(\mathbf{x}, \mathbf{y}) = \sum_{n=0}^{\infty} a_n K(\mathbf{x}, \mathbf{y})^n$. Thus we have by linearity $\nabla_{\mathbf{x}} K_{\text{comp}}(\mathbf{x}, \mathbf{y}) = \sum_{n=0}^{\infty} a_n \nabla_{\mathbf{x}} (K(\mathbf{x}, \mathbf{y})^n) = \sum_{n=0}^{\infty} n a_n K(\mathbf{x}, \mathbf{y})^{n-1} \nabla_{\mathbf{x}} K(\mathbf{x}, \mathbf{y})$ i.e $\|\nabla_{\mathbf{x}} K_{\text{comp}}(\mathbf{x}, \mathbf{y})\| \leq \|\nabla_{\mathbf{x}} K(\mathbf{x}, \mathbf{y})\| \sum_{n=0}^{\infty} n a_n C_K^{n-1} \leq L_K f'(C_K)$. Similarly we have $\sup_{\mathbf{x},\mathbf{y}\in\Omega} \|\nabla_{\mathbf{y}} K_{\text{co}}(\mathbf{x}, \mathbf{y})\| \leq L_K f'(C_K)$. □

We next move on to the Lipschitz properties of $\mathbf{Z}$. Since we have only made assumptions on the expected Lipschtiz properties of $W$, we would only be able to give guarantees on the expected Lipschitz properties of $\mathbf{Z}$. However, as we shall see, these would be sufficient to provide a uniform convergence guarantee over the entire domain $\Omega$. As before, we find that by linearity of expectation, analyzing the expected Lipschitz properties of a single feature map $Z$ are sufficient to guarantee, on expectation, similar properties for $\mathbf{Z}$ as well.

**Lemma 15.** *We have*

$$\sup_{\mathbf{x},\mathbf{y}\in\Omega} \|\nabla_{\mathbf{x}} (Z(\mathbf{x})Z(\mathbf{y}))\| \leq L_W p^2 \sqrt{C_W} f'(pC_W)$$
$$\sup_{\mathbf{x},\mathbf{y}\in\Omega} \|\nabla_{\mathbf{y}} (Z(\mathbf{x})Z(\mathbf{y}))\| \leq L_W p^2 \sqrt{C_W} f'(pC_W)$$

*Proof.* Since $Z(\mathbf{x})Z(\mathbf{y}) = a_N p^{N+1} \prod_{j=1}^{N} W_j(\mathbf{x}) W_j(\mathbf{y})$, by linearity we can write $\nabla_{\mathbf{x}} Z(\mathbf{x})Z(\mathbf{y}) = \left(a_N p^{N+1} \prod_{j=1}^{N} W_j(\mathbf{y})\right) \sum_{j=1}^{N} \left(\prod_{i\neq j} W_i(\mathbf{x})\right) \nabla_{\mathbf{x}} W_j(\mathbf{x})$. Thus we can then write $\|\nabla_{\mathbf{x}} Z(\mathbf{x})Z(\mathbf{y})\|$ as $a_N p^{N+1} \left|\prod_{j=1}^{N} W_j(\mathbf{y})\right| \left\|\sum_{j=1}^{N} \left(\prod_{i\neq j} W_i(\mathbf{x})\right) \nabla_{\mathbf{x}} W_j(\mathbf{x})\right\|$

$$\leq \quad a_N p^{N+1} C_W^{\frac{N}{2}} \sum_{j=1}^{N} C_W^{\frac{N-1}{2}} \|\nabla_{\mathbf{x}} W_j(\mathbf{x})\|$$

which gives us, by linearity of expectation and the bound on the expected Lipschitz properties of the individual estimators,

$$\begin{aligned}
\mathbb{E}\left[\|\nabla_{\mathbf{x}} Z(\mathbf{x}) Z(\mathbf{y})\|\right] &\leq N a_N p^{N+1} C_W^{N-\frac{1}{2}} L_W \\
&= L_W p^2 \sqrt{C_W} \cdot N a_N (p C_W)^{N-1} \\
&\leq L_W p^2 \sqrt{C_W} f'(p C_W)
\end{aligned}$$

Similarly we have $\sup_{\mathbf{x},\mathbf{y} \in \Omega} \|\nabla_{\mathbf{y}} (Z(\mathbf{x}) Z(\mathbf{y}))\| \leq L_W p^2 \sqrt{C_W} f'(p C_W)$. □

Working as before we find that the error function $\mathcal{E}(\mathbf{x}, \mathbf{y}) = \langle \mathbf{Z}(\mathbf{x}), \mathbf{Z}(\mathbf{y})\rangle - K_{\text{co}}(\mathbf{x}, \mathbf{y})$ is, on expectation, $L_1$-Lipschitz for $L_1 = L_K f'(C_K) + L_W p^2 \sqrt{C_W} f'(p C_W)$. Hence the probability that the error function will not be $\frac{\epsilon}{2r}$-Lipschitz is less than $\frac{2L_1 r}{\epsilon}$ by an application of Markov's inequality. However if this is not the case then constructing an $\epsilon$-net at scale $r$ over the domain $\Omega$ and ensuring that the estimator provides an $\epsilon/2$-approximation at centers of these points would ensure an $\epsilon$-accurate estimation to the kernel on the entire domain $\Omega$. Setting up such a net would require at most $\left(\frac{4R}{r}\right)^d$ centers if $\Omega \subseteq \mathcal{B}_1(\mathbf{0}, R)$. Adding the failure probabilities of the estimator not being accurate on the $\epsilon$-net centers to the probability of the error function not being Lipschitz gives us the total error probability of our estimator giving an inaccurate estimate over any point in the domain as $2\left(\frac{4R}{r}\right)^d \exp\left(-\frac{D\epsilon^2}{8C_1^2}\right) + \frac{2L_1 r}{\epsilon}$.

Looking at this quantity as of the form $k_1 r^{-d} + k_2 r$ and setting $r = \left(\frac{k_1}{k_2}\right)^{\frac{1}{d+1}}$ gives us the error probability as $2 k_1^{\frac{1}{d+1}} k_2^{\frac{d}{d+1}} \leq \left(\frac{32 R L_1}{\epsilon}\right) \exp\left(-\frac{D\epsilon^2}{8C_1^2 d}\right)$ if $\epsilon < 8 R L_1$ which gives us the following theorem.

**Theorem 16.** *Let $\Omega \subseteq \mathcal{B}_1(\mathbf{0}, R)$ be a compact subset of $\mathbb{R}^d$ and $K_{co}(\mathbf{x}, \mathbf{y}) = K_{dp}(K(\mathbf{x}, \mathbf{y}))$ be a compositional kernel defined on $\Omega$ satisfying the necessary boundedness and Lipschitz conditions. Assuming we have black-box access to a feature map selection algorithm for $K$ also satisfying the necessary boundedness and Lipschitz conditions, for the feature map $\mathbf{Z}$ defined in Algorithm 2, we have $\mathbb{P}\left[\sup_{\mathbf{x},\mathbf{y} \in \Omega} |\langle \mathbf{Z}(\mathbf{x}), \mathbf{Z}(\mathbf{y})\rangle - K_{co}(\mathbf{x}, \mathbf{y})| > \epsilon\right] \leq \left(\frac{32 R L_1}{\epsilon}\right) \exp\left(-\frac{D\epsilon^2}{8C_1^2 d}\right)$ where $C_1 = p f(p C_W)$ and $L_1 = L_K f'(C_K) + L_W p^2 \sqrt{C_W} f'(p C_W)$ for some small constant $p > 1$. Moreover, with $D = \Omega\left(\frac{d C_1^2}{\epsilon^2} \log\left(\frac{R L_1}{\epsilon \delta}\right)\right)$, one can ensure the same with probability greater than $1 - \delta$.*

Yet again the dependence on input space parameters is similar to that in the case of dot product kernel feature maps. The only non-logarithmic kernel specific dependence is on $C_1$ which encodes the largest possible value taken by the oracle features which is related to the range of values taken by the kernel $K$.

## 6 Experiments

In this section we report results of our feature map construction algorithm on both toy as well as benchmark datasets. In the following, homogeneous kernel refers to the kernel $K_h(\mathbf{x}, \mathbf{y}) = \langle \mathbf{x}, \mathbf{y}\rangle^p$, polynomial kernel refers to $K_p(\mathbf{x}, \mathbf{y}) = (1 + \langle \mathbf{x}, \mathbf{y}\rangle)^p$ and exponential kernel refers to $K_e(\mathbf{x}, \mathbf{y}) = \exp\left(\frac{\langle \mathbf{x}, \mathbf{y}\rangle}{\sigma^2}\right)$. In all our experiments we used $p = 10$ and set the value of the "width" parameter $\sigma$ to be the mean of all pairwise training data distances, a standard heuristic. We shall denote by $d$ the dimensionality of the original feature space and $D$ to be the number of random feature maps used. Before we move on, we describe a heuristic which when used in conjunction with random feature maps gives attractive results allowing for accelerated training and testing times for the SVM algorithm.

### 6.1 The Heuristic H0/1

Consider a dot product kernel defined by $K(\mathbf{x}, \mathbf{y}) = \sum_{n=0}^{\infty} a_n \langle \mathbf{x}, \mathbf{y}\rangle^n$. This heuristic simply makes an observation that the first two terms of this expansion need not be estimated at all. The first term, being a constant, can be absorbed into the offset parameter of SVM formulations and the second term can be handled by simply adjoining the random features with the original features. This allows us to use all our randomness in estimating higher order terms. We refer to algorithmic formulations that use this heuristic as **H0/1** and those that use only random features as **RF**.

We note some properties of this heuristic. First of all, as we shall see, **H0/1** offers superior accuracies even when using a very small number of random features since we get away with an exact estimate of the leading terms in the Maclaurin expansion. However this is accompanied by two overheads. First of all this offers a small overhead while testing since the test vectors are $(d + D)$-dimensional instead of $D$-dimensional if we were to use only random features (as is the case with **RF**).

A more subtle overhead comes at feature map application time since the use of **H0/1** implies that, on an average, each of the $D$ feature maps is estimating a higher order term (as compared to **RF**) which requires more randomness. Moreover, as it takes longer

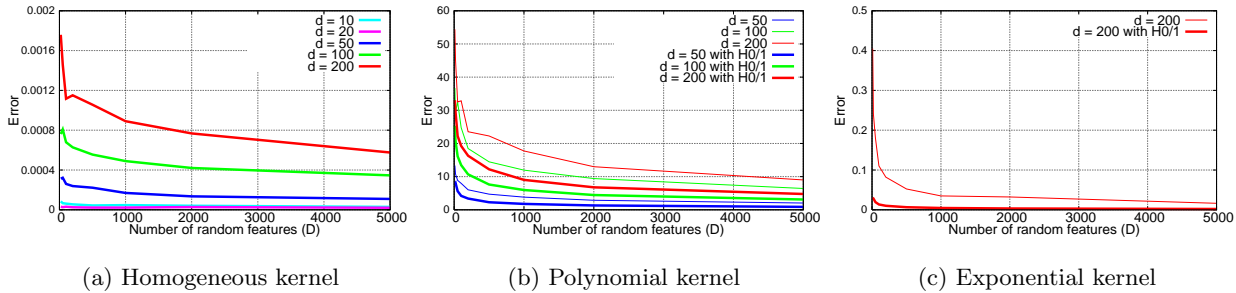

(a) Homogeneous kernel  (b) Polynomial kernel  (c) Exponential kernel

Figure 1: Error rates achieved by random feature maps on three dot product kernels. Plots of different colors represent various values of input dimension $d$. In Figures 1b and 1c, thin plots represent non-**H0/1** experiments and thick plots of same color represent results for the same value of input dimension $d$ but with **H0/1**.

| Dataset | K + LIBSVM | RF + LIBLINEAR | H0/1 + LIBLINEAR | Dataset | K + LIBSVM | RF + LIBLINEAR | H0/1 + LIBLINEAR |
|---|---|---|---|---|---|---|---|
| Nursery<br>N = 13000<br>d = 8 | acc = 99.9%<br>trn = 18.6s<br>tst = 3.37s | acc = 99.7%<br>trn = 3.96s (4.7×)<br>tst = 0.63s (5.3×)<br>D = 500 | acc = 98.2%<br>trn = 0.49s (38×)<br>tst = 0.1s (33×)<br>D = 100 | Nursery<br>N = 13000<br>d = 8 | acc = 99.8%<br>trn = 10.8s<br>tst = 1.7s | acc = 99.6%<br>trn = 2.52s (4.3×)<br>tst = 0.6s (2.8×)<br>D = 500 | acc = 97.96%<br>trn = 0.4s (27×)<br>tst = 0.18s (9.4×)<br>D = 100 |
| Spambase<br>N = 4600<br>d = 57 | acc = 93.8%<br>trn = 3.64s<br>tst = 2.84s | acc = 93.2%<br>trn = 1.67s (2.2×)<br>tst = 1.13s (2.5×)<br>D = 500 | acc = 92.02%<br>trn = 0.19s (19×)<br>tst = 0.38s (7.5×)<br>D = 50 | Spambase<br>N = 4600<br>d = 57 | acc = 93.5%<br>trn = 3.19s<br>tst = 1.89s | acc = 92.3%<br>trn = 1.9s (1.7×)<br>tst = 0.6s (3.1×)<br>D = 500 | acc = 92.08%<br>trn = 0.19s (17×)<br>tst = 0.16s (74×)<br>D = 50 |
| Cod-RNA<br>N = 60000<br>d = 8 | acc = 95.2%<br>trn = 144.1s<br>tst = 28.6s | acc = 94.9%<br>trn = 12.1s (12×)<br>tst = 2.8s (10×)<br>D = 500 | acc = 93.77%<br>trn = 0.63s (229×)<br>tst = 0.51s (56×)<br>D = 50 | Cod-RNA<br>N = 60000<br>d = 8 | acc = 95.2%<br>trn = 91.5s<br>tst = 17.1s | acc = 94.9%<br>trn = 11.5s (8×)<br>tst = 2.8s (6.1×)<br>D = 500 | acc = 93.8%<br>trn = 0.67s (136×)<br>tst = 1.4s (12×)<br>D = 50 |
| Adult<br>N = 49000<br>d = 123 | acc = 84.2%<br>trn = 179.6s<br>tst = 60.6s | acc = 84.7%<br>trn = 21.2s (8.5×)<br>tst = 15.6s (3.9×)<br>D = 500 | acc = 84.7%<br>trn = 6.9s (26×)<br>tst = 7.26s (8.4×)<br>D = 100 | Adult<br>N = 49000<br>d = 123 | acc = 83.7%<br>trn = 263.3s<br>tst = 33.4s | acc = 82.9%<br>trn = 39.8s (6.6×)<br>tst = 14.3s (2.3×)<br>D = 500 | acc = 84.8%<br>trn = 7.18s (37×)<br>tst = 9.4s (3.6×)<br>D = 100 |
| IJCNN<br>N=141000<br>d = 22 | acc = 98.4%<br>trn = 164.1s<br>tst = 33.4s | acc = 97.3%<br>trn = 36.5s (4.5×)<br>tst = 23.3s (1.4×)<br>D = 1000 | acc = 92.3%<br>trn= 4.98s (33×)<br>tst = 7.5s (4.5×)<br>D = 200 | IJCNN<br>N=141000<br>d = 22 | acc = 98.4%<br>trn = 135.8s<br>tst = 29.98s | acc = 97.2%<br>trn = 24.9s (5.5×)<br>tst = 23.4s (1.3×)<br>D = 1000 | acc = 92.2%<br>trn = 5.2s (26×)<br>tst = 9.1s (3.3×)<br>D = 200 |
| Covertype<br>N=581000<br>d = 54 | acc = 77.4%<br>trn = 160.95s<br>tst = 1653.9s | acc = 77.04%<br>trn = 186.1s (—)<br>tst = 236.8s (7×)<br>D = 1000 | acc = 75.5%<br>trn = 3.9s (41×)<br>tst = 70.3s (23×)<br>D = 100 | Covertype<br>N=581000<br>d = 54 | acc = 80.6%<br>trn = 194.1s<br>tst = 695.8s | acc = 76.2%<br>trn = 21.4s (9×)<br>tst = 207s (3.6×)<br>D = 1000 | acc = 75.5%<br>trn = 3.7s (52×)<br>tst = 80.4s (8.7×)<br>D = 100 |

(a) Polynomial Kernel, $K(\mathbf{x}, \mathbf{y}) = (1 + \langle \mathbf{x}, \mathbf{y} \rangle)^{10}$     (b) Exponential Kernel, $K(\mathbf{x}, \mathbf{y}) = \exp\left(\frac{\langle \mathbf{x}, \mathbf{y} \rangle}{\sigma^2}\right)$ for feature maps estimating higher order terms to be applied (see Algorithm 1), this results in longer feature construction times. Hence, after $D$ is chosen beyond a certain threshold, the benefits offered by **H0/1** are overshadowed by the longer feature construction times and plain **RF** becomes more preferable in terms of lower test times. However, as the experiments will indicate, **H0/1** is an attractive option for ultra fast learning routines for small to moderate values of $D$ which, although do not increase feature construction time too much, offer much better classification accuracies than **RF**.

Table 1: **RF**, **H0/1** and **K** denote respectively, the use of random features, **H0/1** and actual kernel values. The first columns list the datasets, their sizes (N) and their dimensionalities (d). Subsequent columns list the number of random features used (D), classification accuracies (acc), training/testing times (trn/tst) and speedups (×).

### 6.2 Toy Experiments

In our first experiment, we tested the accuracy of the feature maps on the three dot product kernels $K_h$, $K_p$ and $K_e$. We sampled 100 random points from the unit ball in $d$ dimensions (we used various values of $d$ between 10 and 200) and constructed feature maps for various values of $D$ from 10 to 5000. The error incurred by the feature maps was taken to be the average absolute difference between the entries of the kernel matrix as given by the dot product kernel and that given by the linear kernel on the new feature space given by the feature maps. The results of the experiments, averaged over 5 runs are shown in Figure 1. One can see that in each case, the error quickly drops as we increase the value of $D$.

We also experimented with the effect of **H0/1** on these toy datasets for $K_p$ and $K_e$ ($K_h$ does not have terms corresponding to $n = 0, 1$ and hence **H0/1** cannot be applied). For sake of clarity, the X-axis in all the graphs in Figure 1 represent only $D$ and not the final number of features used (which is $d + D$ for **H0/1** experiments). Also, to avoid clutter, we have omitted plots for certain small values of $d$ in Figures 1b and 1c. Notice how in all cases, **H0/1** registers a sharper drop in error than **RF**.

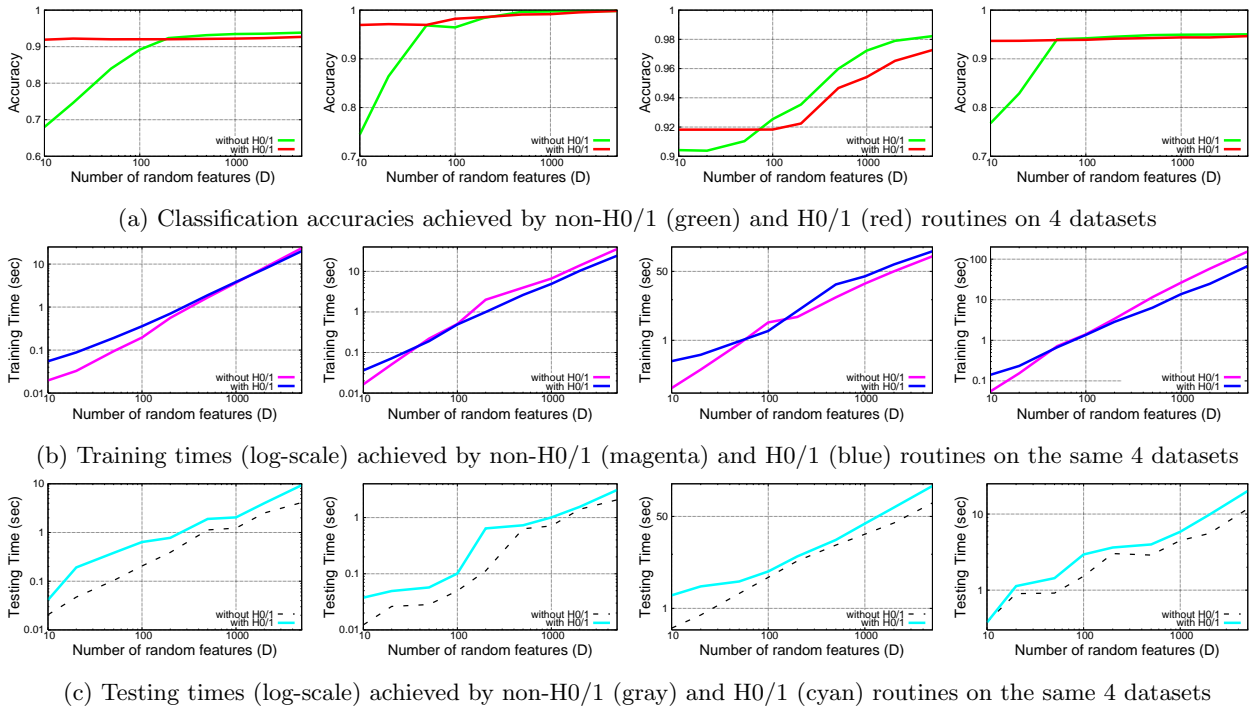

Figure 2: Performance of **H0/1** vs non-**H0/1** on four datasets. The first column corresponds to experiments on the `Spambase` dataset with the polynomial kernel. The next three columns correspond to experiments on `Nursery` with the polynomial kernel, `IJCNN` with the exponential kernel and `Cod-RNA` with the exponential kernel.

We note that the error rates vary considerably across kernels. This is due to the difference in the range of values taken by these kernels. With the specified values of kernel parameters, whereas $K_h$ can only take values in the range $[-1, 1]$ inside $\mathcal{B}_2(\mathbf{0}, 1) \subset \mathbb{R}^d$, $K_p$ can take values up to 1024 and $K_e$ up to 2.73. One notices that the error rates offered by the feature maps also differ in much the same way for these kernels.

### 6.3 Experiments on UCI Datasets

In our second experiment, we tested the performance of our feature map on benchmark datasets. In these experiments we used 60% of the data (subject to a maximum of 20000) for training and the rest as test data. Non-linear kernels were used alongwith LIB-SVM (Chang and Lin, 2011) and random feature routines **RF** and **H0/1** were used alongwith LIBLINEAR (Fan et al., 2008) for the classification tasks. Non-binary problems were binarized randomly for simplicity. Since the kernels being considered are unbounded, the lengths of all vectors were normalized using normalization constants learnt on the training sets. All results presented are averages across five random (but fixed) splits of the datasets.

We first take a look at the performance benefits of **H0/1** on these datasets in Figure 2. As before we simply plot $D$ on the X-axis even for **H0/1** experiments for sake of clarity. We observe that in all four cases, **H0/1** offers much higher accuracies as compared to **RF** when used with small number of random features (see Figure 2a). Also note that the number of extra features added for **H0/1** is not large (avg. $d = 45$ for the 6 datasets considered). As we increase the number of random features, **H0/1** accuracies move up slowly. However the test feature construction overhead become large after a point and affects test times (see Figure 2c). The effect on training times (see Figure 2b) is not so clear since the use of **H0/1** also seems to offer greater separability which mitigates the training feature construction overhead in some cases.

We provide details of the results in Table 1. We see that both **RF** and **H0/1** offer significant speedups in both training and test times while offering competitive classification accuracies with **H0/1** doing so at much lower values of $D$. In some cases the reduction in classification accuracy for **H0/1** is moderate but is almost always accompanied with a spectacular increase in training and test speeds.

**Acknowledgements**

The authors thank the anonymous referees for comments that improved the presentation of the paper. P. K. thanks Prateek Jain and Manik Varma for useful discussions and Devanshu Bhimwal for pointing out an error in Lemma 10. P. K. is supported by Microsoft Corporation and Microsoft Research India under a Microsoft Research India Ph.D. fellowship award.